\documentclass{article}
\usepackage{ijcai21}

\usepackage{times}
\usepackage{soul}
\usepackage{url}
\usepackage{color}
\usepackage[hidelinks]{hyperref}
\usepackage[utf8]{inputenc}
\usepackage[small]{caption}
\usepackage{graphicx}
\usepackage{amsmath}
\usepackage{amsthm}
\usepackage{amsfonts}
\usepackage{booktabs}
\usepackage[ruled,linesnumbered]{algorithm2e}
\usepackage{subfigure}
\usepackage{float}
\usepackage{bbm}
\usepackage{xcolor}
\usepackage[switch]{lineno}

\title{Direct Measure Matching for Crowd Counting}

\author{
   Hui Lin$^1$\and Xiaopeng Hong$^{1,4}$\footnote{Corresponding author}\and Zhiheng Ma$^2$\and Xing Wei$^3$\and Yunfeng Qiu$^3$\and Yaowei Wang$^4$\and Yihong Gong$^3$
    \affiliations
    $^1$School of Cyber Science and Engineering, Xi'an Jiaotong University;\\
    $^2$College of Artificial Intelligence, Xi'an Jiaotong University;\\
    $^3$School of Software Engineering, Xi'an Jiaotong University;\\
    $^4$Pengcheng Laboratory, Shenzhen
    \emails
    waitandwait@stu.xjtu.edu.cn; hongxiaopeng@ieee.org;
    mazhiheng@stu.xjtu.edu.cn;
    weixing@mail.xjtu.edu.cn;
    yfqiu2015@stu.xjtu.edu.cn;
    wangyw@pcl.ac.cn;
    ygong@mail.xjtu.edu.cn
}

\begin{document}

\maketitle

\begin{abstract}
  {Traditional crowd counting approaches usually use Gaussian assumption to generate pseudo density ground truth, which suffers from problems like inaccurate estimation of the Gaussian kernel sizes. In this paper, we propose a new measure-based counting approach to regress the predicted density maps to the scattered point-annotated ground truth directly. First, crowd counting is formulated as a measure matching problem. Second, we derive a semi-balanced form of Sinkhorn divergence, based on which a Sinkhorn counting loss is designed for measure matching. Third, we propose a self-supervised mechanism by devising a Sinkhorn scale consistency loss to resist scale changes. Finally, an efficient optimization method is provided to minimize the overall loss function. Extensive experiments on four challenging crowd counting datasets namely ShanghaiTech, UCF-QNRF, JHU++ and NWPU have validated the proposed method.}
  
\end{abstract}

\section{Introduction}

Crowd counting has become increasingly important in the fields of artificial intelligence and computer vision. It has been widely used in congestion estimation and crowd management. With the use of Convolutional Neural Networks (CNNs), crowd counting has achieved considerable success in recent years. However, due to complex images and coarse (point) annotations, the task is still challenging.

Existing crowd counting methods can be briefly categorized into two types. Detection based methods count the number of people by exhaustively detecting every individual in images~\cite{liu2019point}~\cite{liu2018decidenet}. Their applications are limited as they usually require additional annotations such as bounding boxes. Regression based methods regress the output to a \emph{pseudo} density map by smoothing scattered annotated points with a fixed-size Gaussian kernel~\cite{zhang2016single}~\cite{sindagi2017generating}~\cite{zhang2018crowd}.
However, the quality of the generated pseudo maps highly depends on the settings of the Gaussian kernel size. It is shown that inappropriate sizes of Gaussian kernels will greatly impair the quality of density maps~\cite{idrees2018composition}.

\begin{figure}[t]
\setlength{\abovecaptionskip}{0cm}
\begin{center}
    \includegraphics[width=0.47\textwidth]{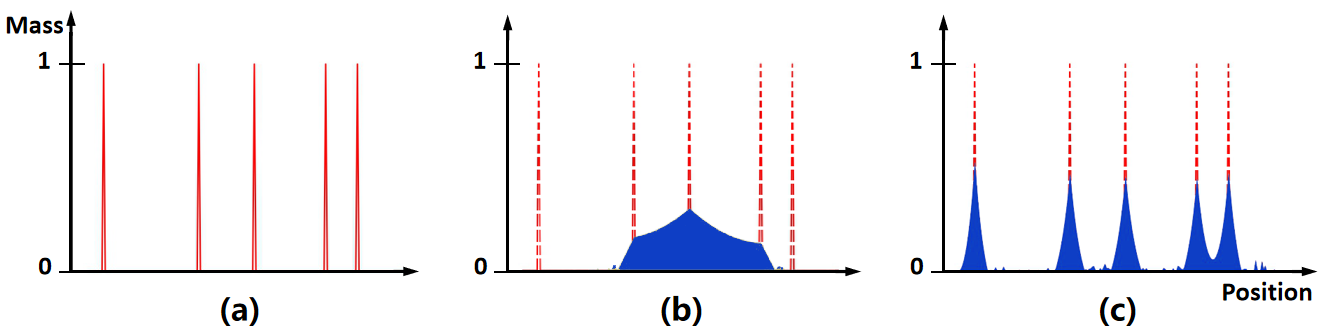}
\end{center}
\caption{{Comparisons of using Wasserstein distance (WD) and our proposed semi-balanced Sinkhorn divergence (SSD) as regressive loss. (a): The regression target. (b): The WD based regression results may shrink to a mass and suffers from \emph{entropic bias}. (c): The SSD based regression output is sharp, well separated, and clearly centered at the annotated points.}}
\label{fig:entro}
\end{figure}


{To solve these problems, in this paper, we regard crowd counting as a measure matching problem, based on the understanding that the scattered ground truth and the predicted density map can be expressed by a discrete point measure and a continuous measure, respectively.}




{It is essential to choose appropriate distance metric when matching two measures of different types. This may at first glance appear trivial as regularized Wasserstein distance is recently popular to calculate the discrepancy between measures. The regularized Wasserstein distance usually appends an entropic function as a regularization term to the overall objective function to relax the hard constraints during the assignment of optimal transport amount and reduce computational costs~\cite{cuturi2013sinkhorn}}. However, {there} exist two serious problems. First, the entropic term breaks the {the axiom of \emph{Identity of indiscernibles}, i.e., $d(x,x) = 0$, which is the fundamental property of a metric space. Second, the quality of the optimization is severely sensitive to the parameter settings of the entropic term~\cite{feydy2019interpolating}. Inappropriate entropic parameters will cause the trainable measure shrink to a mass at the barycenter of the target. This phenomenon is often referred to as \emph{entropic bias}, which is visualized in Figure~\ref{fig:entro}.}

{Sinkhorn divergence can be an opinion to fix \emph{the identity of indiscernibles} and the \emph{hard constraints issues}. Sinkhorn divergence is with a self-correcting term. Thus it is more stable under different entropic parameters and with a better interpolation meaning, compared to Wasserstein distance~\cite{ramdas2017wasserstein}.}
Nevertheless, it is still constrained by the {\emph{equivalence of measures' quantity}, which requires that the total amount of predicted density measure equals to the amount of ground truth measure}. {As a result, it cannot be directly applied to practical scenarios where the total mass of the predicted density map is different to those of the scattered annotation maps.} 

\begin{figure}[t]
\begin{center}
    \includegraphics[width=0.47\textwidth]{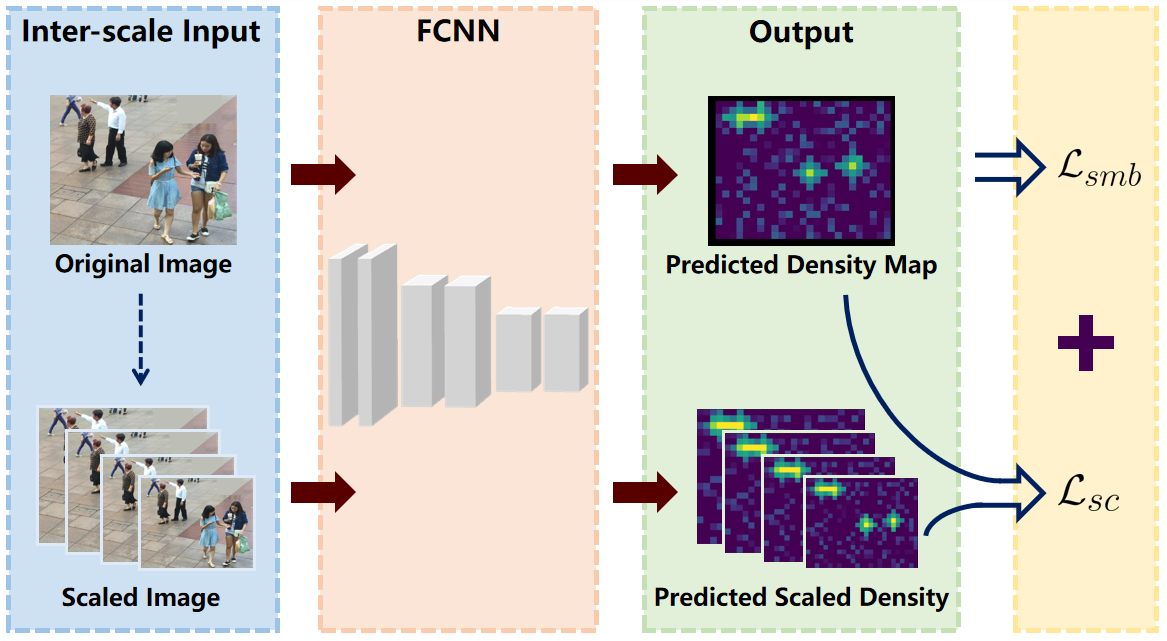}
\end{center}
\caption{The framework of our proposed method S3. It has two compositions. First, the counting loss $\mathcal{L}_{smb}$ directly measures the divergence from ground truth by the semi-balanced Sinkhorn. Second, the scale consistency loss $\mathcal{L}_{sc}$ punishes the deviations in the measure space of various crowd scales.}
\label{fig:structure}
\end{figure}

{In this paper, we  propose a new measure-theory based counting approach which directly regresses to point annotations, {termed by semi-balanced Sinkhorn with scale consistency (S3)}.
Firstly, to break the aforementioned limitations, we derive a semi-balanced form of the Sinkhorn distance and design a \emph{semi-balanced Sinkhorn counting loss}. This new formulation relaxes the amount constraint and is fully in line with our problem assumption.
Secondly, although the information about crowd scales has shown to be significant to counting {~\cite{zhang2016single}~\cite{zeng2017multi}~\cite{cao2018scale}}, it is lost when only the point annotations are available. To overcome this deficiency, we propose a self-supervised scale enhanced mechanism by using the inter-scale consistency in Sinkhorn distance and devise a \emph{Sinkhorn scale consistency loss}. Thirdly, we derive the first-order conditions of the overall loss function and introduce the scaling iterations for efficient optimization.}
The framework of our method is depicted in Figure~\ref{fig:structure}. 


To evaluate our method, we have conducted extensive experiments and achieved {promising results. The superior performance of our method is also demonstrated from the perspective of visualizations, where the outputs of the proposed method appear sharp and locate closed to annotated points.}




The contributions of this paper are summarized as follows:
\begin{itemize}

\item {We propose a new measure-matching based crowd counting approach, which can directly regress the dense density map to the scattered point annotations, without using Gaussian assumptions to generate poor-quality \emph{pseudo} ground truth.}

\item  {We derive a semi-balanced form of Sinkhorn divergence for computing the distance between two heterogeneous measures with different masses.} 

\item {We propose a self-supervised scale enhanced mechanism to improve the robustness against scale variations.}






\item Extensive experiments illustrate that our method achieves highly competitive counting performance.
\end{itemize}

\section{Related Works}

\subsection{Crowd Counting}

Crowd Counting has experienced rapid development due to the supports of a myriad of methods. Early papers tended to adopt detection of heads or bodies of crowd~\cite{wu2005detection} but are limited by the high-density crowd congestion. Some works have introduced direct regressions with low-level features~\cite{chen2012feature}~\cite{brostow2006unsupervised}.
{More recently, deep CNN based crowd counting methods have achieved outstanding performances.} \cite{zhang2016single} presents a multi-column CNN which regresses to the pseudo density map with adaptive Gaussian kernels.~\cite{li2018csrnet} applies a dilated network suitable for highly congested scenes.~\cite{zhang2019relational} proposes a relational attention network with exploration of interdependence of pixels. Moreover, methods based on segmentation~\cite{sajid2016crowd}, perspective estimation~\cite{yan2019perspective}, error estimation~\cite{he2021error} and multi-scale mechanisms~\cite{zeng2017multi}~\cite{sindagi2017generating}~\cite{liu2019context} are proposed by latest papers, in order to break the limitations of perspective and scale variations of crowds. Furthermore, the method~\cite{ma2019bayesian} adopts Bayesian assumption and calculates with the expected count of pixels.
{Our method is distinct to most existing approaches. We treat crowd counting as a \emph{measure matching} problem while others usually regard them as \emph{density-map-matching} ones. Moreover, we introduce the measure theory based distance to gauge the distances of measures, which are beyond the scope of most existing crowd counting studies.}

Recently, ~\cite{wang2020distribution} tries to use optimal transport (OT) to measure the similarity between the normalized predicted and ground truth density map. But it is still constrained by the amount of density map and some unpleasing properties of traditional optimal transport.




\subsection{Wasserstein and Sinkhorn Divergences}


{Wasserstein distance, \emph{a.k.a.} Earth Mover's Distance, provides an efficient way of calculating the distance between measures. It has many favorable properties, \emph{e.g.}, convexity, tightness and the existence of optimal couplings~\cite{villani2008optimal}, and thus has been widely used in many artificial intelligence applications these years. Some} generative adversarial networks employ Wasserstein distance to generate computing functions with better properties~\cite{arjovsky2017wasserstein}. It has also been extensively adopted in domain adaptation~\cite{shen2018wasserstein} and metric learning~\cite{xu2018multi}.

Alternatively, Sinkhorn divergence eliminates the entropic bias and gathers the respective strengths between Wasserstein distance and MMD~\cite{ramdas2017wasserstein}~\cite{feydy2019interpolating}. It was early proposed to solve the generative models~\cite{genevay2018learning}.~\cite{sejourne2019sinkhorn} extends it to unbalanced optimal transport and elaborates on its attractive properties. To the best of our knowledge, there are no existing studies on using Sinkhorn distance for calculating between the distributions of two crowds.


\section{The Proposed Method}

In this section, we will detail our measure theory based semi-balanced Sinkhorn divergence with scale consistency. By first defining the problem as measure matching, we then explain the shortages of using traditional Wasserstein and Sinkhorn for evaluating the divergence. We elaborate the proposed semi-balanced Sinkhorn distance and use it as our counting loss. Then, we adopt inter-scale consistency mechanism and measure the deviations as scale loss. We combine these two losses for jointly regression. Finally, we will also investigate on computing solutions. 

\subsection{Problem Definition}

Traditional method, which is based on pseudo density map, regards the counting problem as density regression. It adopts $L_2$ loss by generating the Gaussian-kernel density map with same pixel number as the output.

Let $\mathcal{X}$, $\mathcal{Y}$ denote the 2-D supports of estimated density map and ground truth respectively. By redefining the problem as measure matching, from the perspective of measure theory, we represent the ground truth by $\boldsymbol\beta=\sum_{j=1}^{M}{\beta(y_j)}\delta_{y_j}$. Since $j$ represents an annotated point (a person in crowd counting), $\beta(y_j)=1$, $m(\boldsymbol\beta)=\sum_{j=1}^{M}{\beta(y_j)}$ equals to the number of people $M$. $y_j\in\mathcal{Y}$ signals the location of person $j$, and $\delta_{y_j}$ is a unit Dirac located at $y_j$.

Similar to the point measure, the output of density regressor can be defined as a continuous measure:
\begin{center}
$\boldsymbol\alpha=\int_{\mathcal{X}} \alpha(x)\mathrm{d}x$, where $\alpha(x) = R_{\theta}(x;I)$.
\end{center}
$R_{\theta}$ is the density regressor with the trainable parameter $\theta$ and $\alpha(x)$ is nonnegative.

Consequently, the objective of regression is to reduce the discrepancy between the ground truth measure and the output measure. However, considering that $\mathcal{X}$ and $\mathcal{Y}$ are supports of continuous and discrete measures respectively, it is challenging to calculate the divergence between these two heterogeneous measures which have few overlaps. Therefore, the optimization will become laborious and sometimes impractical. To address this limitation, we seek to propose a new method which can efficiently compute the distance.

\subsection{Wasserstein and Sinkhorn Divergences}

Wasserstein distance has recently been used as an efficient way to associate different measures. It aims to calculate the minimum discrepancy by finding the optimal transport map $\pi(x,y)$, which describes the amount of mass tranporting from output measure to ground truth measure. Using $c(x,y)$ as moving costs, Wasserstein divergence can be expressed by:
\begin{equation}
W(\boldsymbol\alpha,\boldsymbol\beta)=\min_{\pi(x,y)}\ \int_{\mathcal{X}\times\mathcal{Y}}c(x,y)\mathrm{d}\pi(x,y),
\end{equation}
where $\sum_{j=1}^{M}\pi(x,y_j)=\alpha(x)$, $\int_\mathcal{X}\mathrm{d}\pi(x,y_j)=\beta(y_j)$. However, as a linear program, this divergence suffers from a high computational cost. One effective way to release from the burden is to find an approximating version.

Typically, entropic regularization~\cite{cuturi2013sinkhorn} is widely used in recent papers. By defining a regularized expression, the distance changes into:
\begin{equation}
\begin{aligned}
W_\varepsilon(\boldsymbol\alpha,\boldsymbol\beta)=&\min_{\pi(x,y)}\ \int_{\mathcal{X}\times\mathcal{Y}}c(x,y)\mathrm{d}\pi(x,y)\\
&+\varepsilon\int_{\mathcal{X}\times\mathcal{Y}}\ln{(\frac{\pi(x,y)}{{\alpha(x)}{\beta(y)}})}\mathrm{d}\pi(x,y).
\end{aligned}
\end{equation}
Entropic parameter $\varepsilon$, in general, is positive, determining the degree of smoothing. When $\varepsilon \rightarrow 0$, the distance converges to unregularized one. Although the regularization helps to solve the computation problem efficiently, as $\varepsilon$ increases, when trying to optimize $\boldsymbol\alpha$ by minimizing the discrepancy, $\boldsymbol\alpha$ begins to shrink and the deviation expands~\cite{feydy2019interpolating}. It is obvious when $\varepsilon \rightarrow +\infty$, $\boldsymbol\alpha$ will be converged to a Dirac located at the barycenter of static $\boldsymbol\beta$. Meanwhile, as 
\begin{center}
    $W_\varepsilon(\boldsymbol\alpha,\boldsymbol\alpha)\ >\ 0$, when $\varepsilon>0$,
\end{center}
the regularized Wasserstein distance does not satisfy the identity of indiscernibles.


To address this, we extend the Wasserstein distance to Sinkhorn distance~\cite{genevay2018learning}:
\begin{equation}\label{epsilon sinkhorn}
S_\varepsilon(\boldsymbol\alpha,\boldsymbol\beta)=W_\varepsilon(\boldsymbol\alpha,\boldsymbol\beta)-\frac{1}{2}W_\varepsilon(\boldsymbol\alpha,\boldsymbol\alpha)-\ (\frac{1}{2}W_\varepsilon(\boldsymbol\beta,\boldsymbol\beta)),
\end{equation}
where deviation is prevented by a self-correcting term $W_\varepsilon(\boldsymbol\alpha,\boldsymbol\alpha)$. This debiased formula helps to guarantee the identity of indiscernibles, $S_\varepsilon(\boldsymbol\alpha,\boldsymbol\alpha)=0$, so that the entropic bias is eliminated when matching $\boldsymbol\alpha$ to ground truth $\boldsymbol\beta$.

Unfortunately, there is still another limitation. One major requirement of above equations is $\int_\mathcal{X}\mathrm{d}\alpha(x)=\int_{\mathcal{X}\times\mathcal{Y}}{\pi(x,y)}=\sum_{j=1}^{M}{\beta(y_j)}$. As we cannot control the total amount of the predicted density measure, directly using this divergence will violate the requirement and trigger severe mathematical problems. Meanwhile, different from the unbalanced assumption that both measures are uncontrollable, the total amount of ground truth measure has been already known. Therefore, to adapt with our problem, we propose a novel semi-balanced form to break this constraint. Details will be explained in the following sections.


\subsection{Semi-balanced Sinkhorn Divergence}

Here, we give derivations and properties of our semi-balanced Sinkhorn distance. Compared to traditional distances above, it expresses the relationship between two heterogeneous measures more directly. First, we {define} $W_\varepsilon^{smb}$ as follow:
\begin{equation}
\begin{aligned}
&W_\varepsilon^{smb}(\boldsymbol\alpha,\boldsymbol\beta)=\min_{\pi(x,y)}\int_{\mathcal{X}\times\mathcal{Y}}c(x,y)\mathrm{d}\pi(x,y)+\\
&D_\varphi(\sum_{j=1}^{M} {\pi(x,y_j)|\alpha(x)})+\varepsilon\int_{\mathcal{X}\times\mathcal{Y}}\ln{(\frac{\pi(x,y)}{{\alpha(x)}{\beta(y)}})}\mathrm{d}\pi(x,y),
\end{aligned}
\end{equation}
where $D_\varphi$ is a transport penalty which relaxes the strict constraint, allowing the amount of moving mass can be differ to measure $\boldsymbol\alpha$. From an intuitive point of view, the formula gives $\boldsymbol\alpha$ slight value changes while assigning to different points. Given $\mathcal{K}$ is the support of measure $\boldsymbol{\alpha_1}$ and $\boldsymbol{\alpha_2}$, the penalty is defined by:
\begin{equation}
D_{\varphi}({\boldsymbol{\alpha_1}}|{\boldsymbol{\alpha_2}})=\int_\mathcal{K} \varphi(\frac{\alpha_{1k}}{\alpha_{2k}})\ \mathrm{d}\alpha_{2k}.
\end{equation}

To guarantee that $D_\varphi$ is nonnegative and proper, $\varphi$ function should satisfy $\varphi(1)=0$ and be strictly convex. Typically, when $\sum_{j=1}^{M}\pi(x,y_j)=\alpha(x)$, $\varphi(1)=0$, indicating that the formulation degenerates to the classical Wasserstein distance. Detailed properties and the deviation will be further explained in Section~\ref{overall loss and optimization}.

Considering that the self-correcting term $W_\varepsilon(\boldsymbol\alpha,\boldsymbol\alpha)$ is inherently satisfied the amount equivalence, we can directly keep the original balanced form. Thus, we propose the semi-balanced Sinkhorn distance as:
\begin{equation}
\begin{aligned}
S_\varepsilon^{smb}(\boldsymbol\alpha,\boldsymbol\beta)=&W_\varepsilon^{smb}(\boldsymbol\alpha,\boldsymbol\beta)-\frac{1}{2}W_\varepsilon(\boldsymbol\alpha,\boldsymbol\alpha)\\
&{+\frac{\varepsilon^2}{2}(m(\boldsymbol\alpha)-m(\boldsymbol\beta))^{2}.}
\label{sinkhorn}
\end{aligned}
\end{equation}


\subsection{{Scale Consistency at Measure Space}}


{Most existing crowd counting datasets only provide point annotations and there is no scale information available. As a result, we cannot quantify the scale information of the instances explicitly and have to regard the ground truth as Dirac measures. This may harm the counting accuracy as the scale information has turned out to be contributive~~\cite{zhang2016single,zeng2017multi,cao2018scale}.}

Previous methods have tried to address this issue by adopting multi-scale network~\cite{liu2019context}~\cite{sindagi2017generating}~\cite{ma2020learning} or designing external scale modules or losses~\cite{xu2019learn}~\cite{shen2018crowd}. Different from them, in this paper, we propose a self-supervised scale enhanced mechanism from the perspective of measure theory. Our proposed mechanism is able to measure the divergence under different scale pyramid directly without normalizing them to the same sizes. 

First, We resize the original image $I$ as $Sc(I)$, and obtain the scaled output measure $\hat{\boldsymbol\alpha}=R_{\theta}(Sc(I))$. Then, we perform the same re-scale transform on $\boldsymbol\alpha=R_{\theta}(I)$ to get $Sc(\boldsymbol{\alpha})$. In measure space, $\hat{\boldsymbol\alpha}$ and $Sc(\boldsymbol\alpha)$ are expected to be closed, and ideally, $Distance(\hat{\boldsymbol\alpha},Sc(\boldsymbol\alpha))=0$. Otherwise, if a model is sensitive to scale variations, there will be a unignored deviation between $\hat{\boldsymbol\alpha}$ and $Sc(\boldsymbol\alpha)$. To improve the robustness towards scale changes, the key is to minimize the scale measure divergence. This understanding leads us to punish the deviation and design a Sinkhorn scale consistency loss based on measure theory as follows:
\begin{equation}
\mathcal{L}_{sc}= S_\varepsilon(\hat{\boldsymbol\alpha},Sc(\boldsymbol\alpha)).
\label{sc loss}
\end{equation}


\begin{figure*}[t!]
	\begin{center}
		\setlength{\tabcolsep}{1pt}
		\begin{tabular}{ccccc}
			\includegraphics[height=0.13\linewidth]{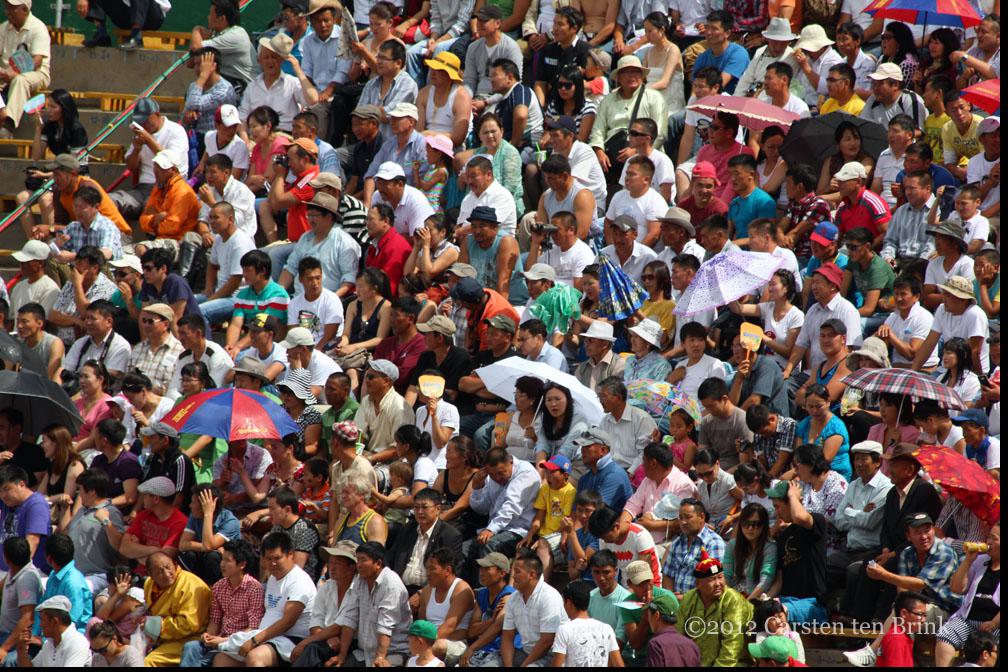}  &
			\includegraphics[height=0.13\linewidth]{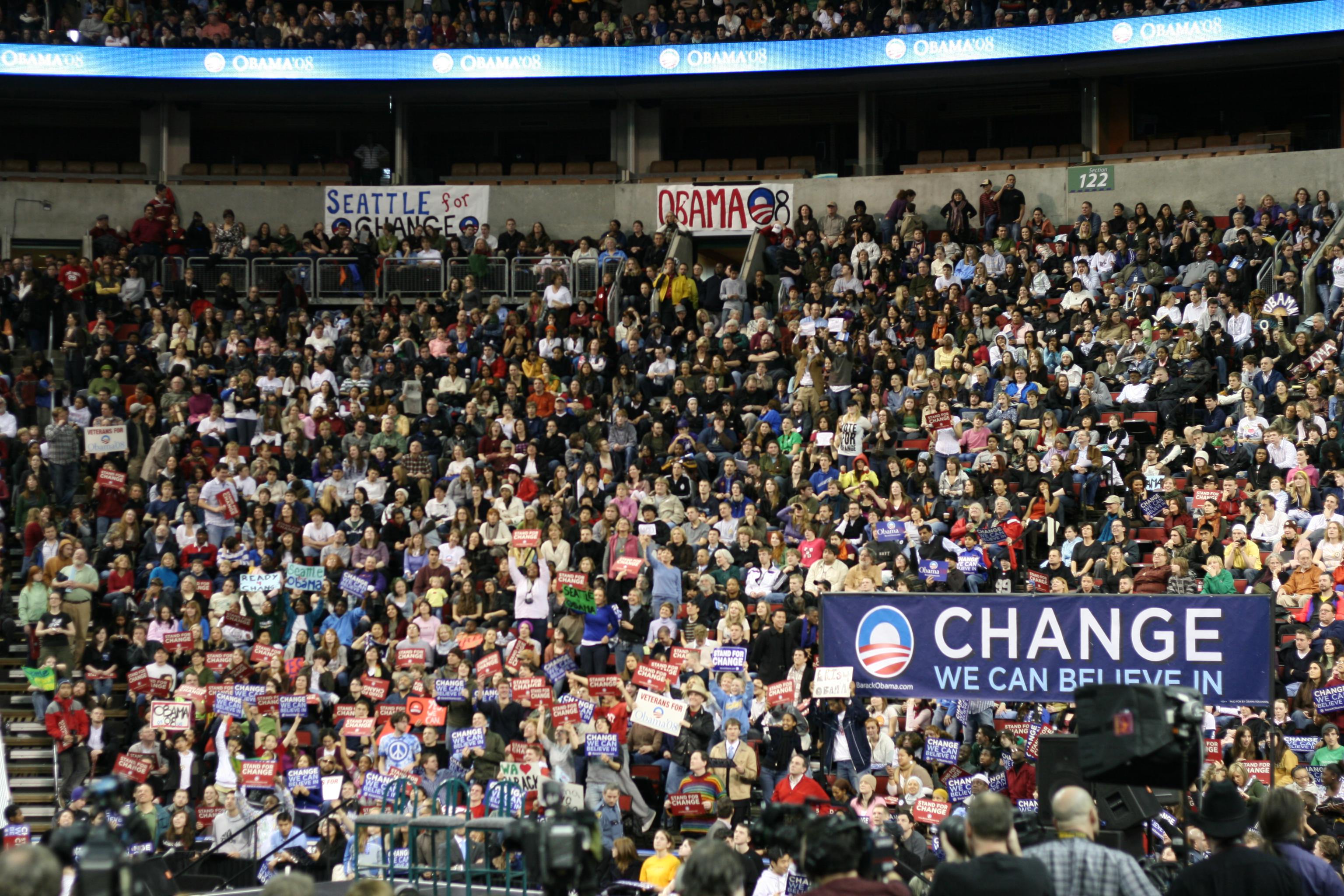}  &
			\includegraphics[height=0.13\linewidth]{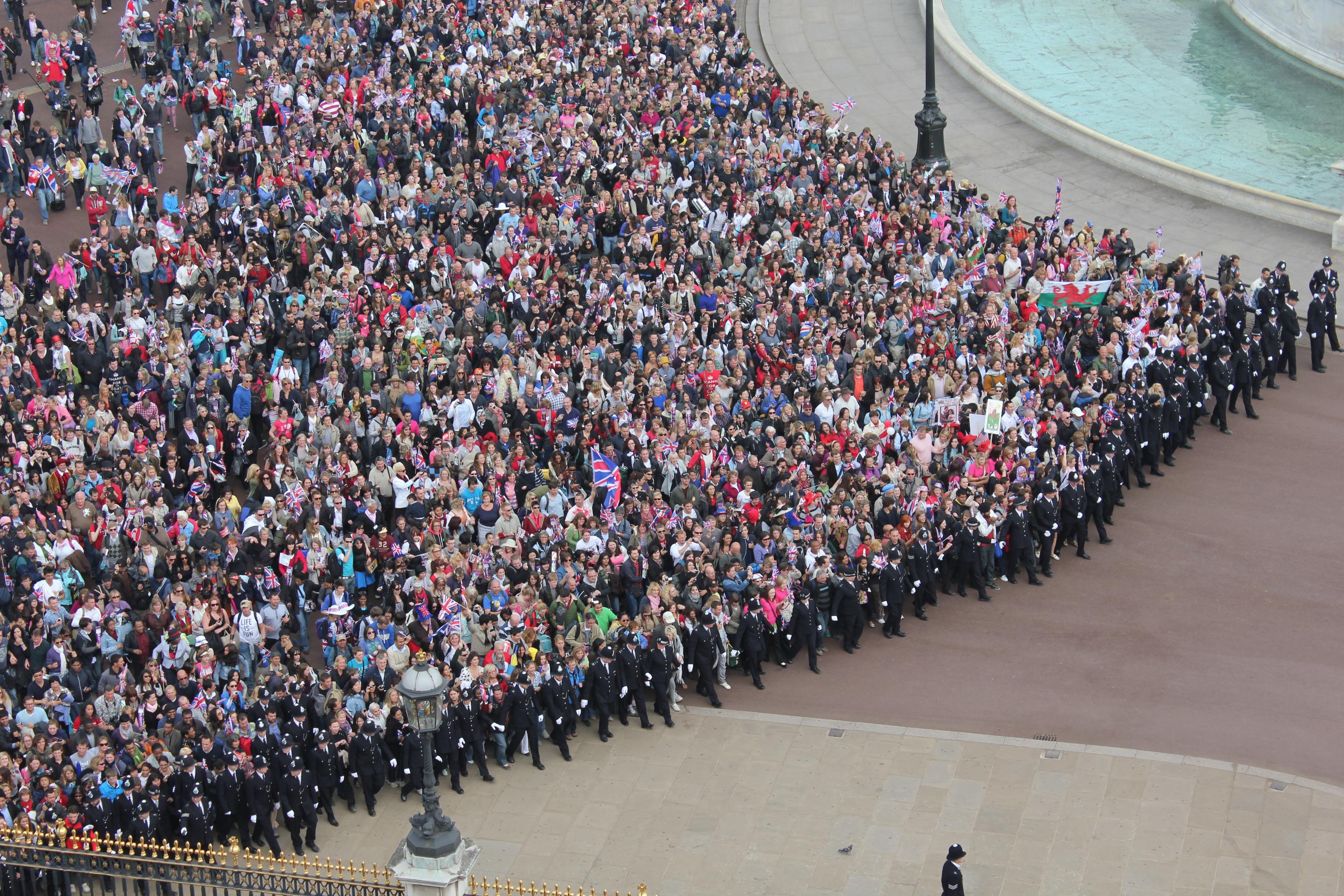}  &
			\includegraphics[height=0.13\linewidth]{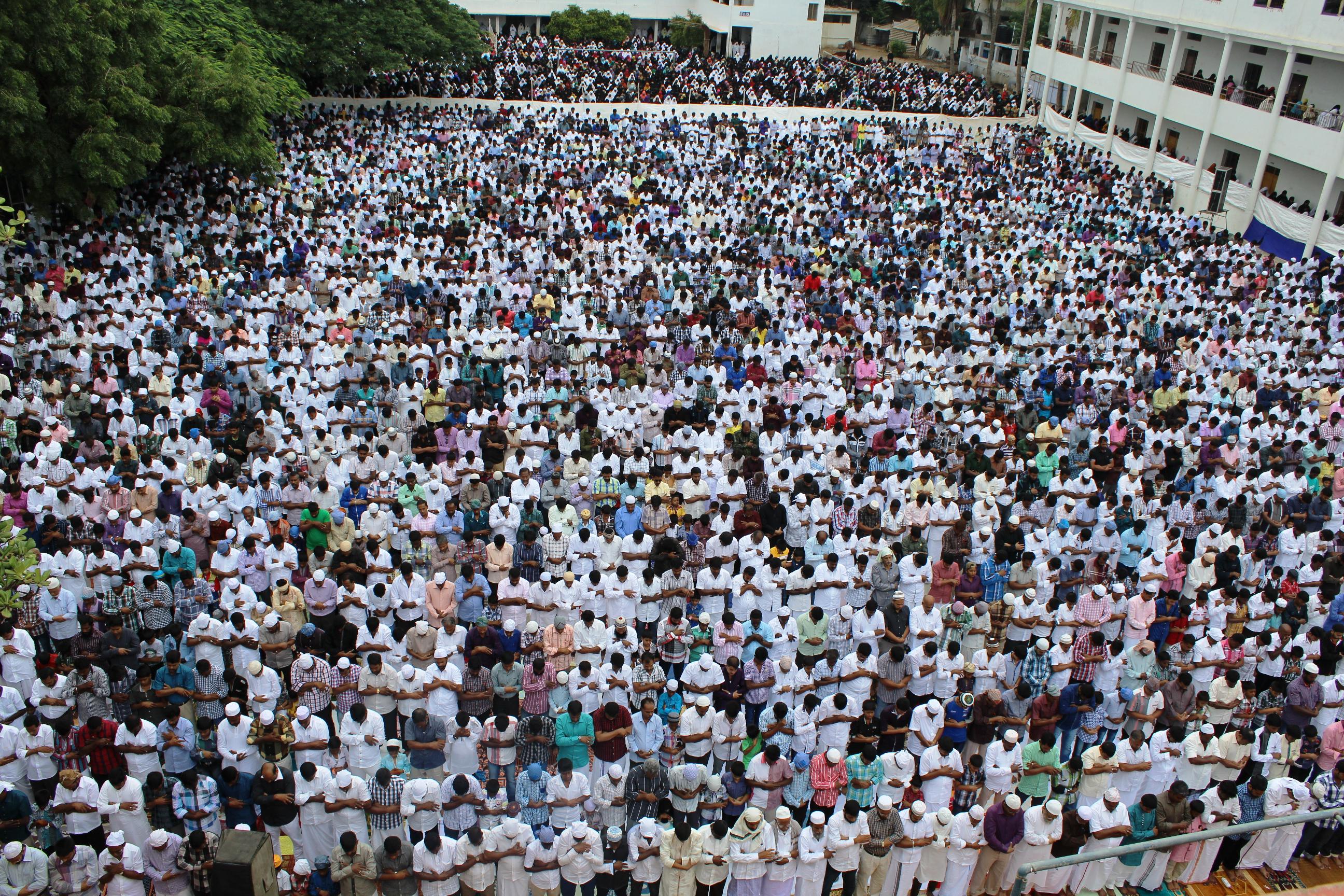}  &
			\includegraphics[height=0.13\linewidth]{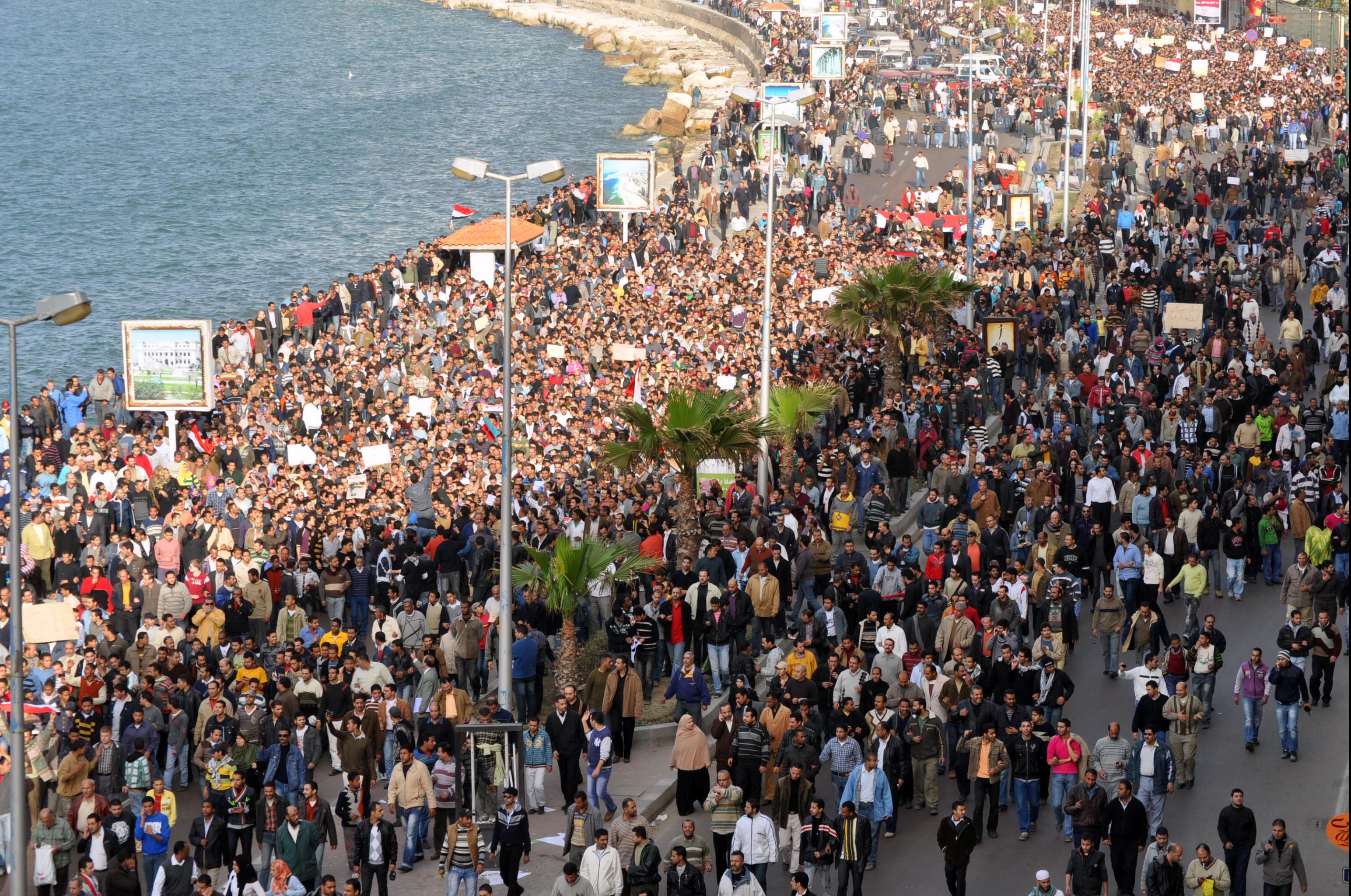}\\
			\footnotesize{gt: 191} & \footnotesize{gt: 909} & \footnotesize{gt: 2165} & \footnotesize{gt: 3970} & \footnotesize{gt: 4195} \\
			\includegraphics[height=0.13\linewidth]{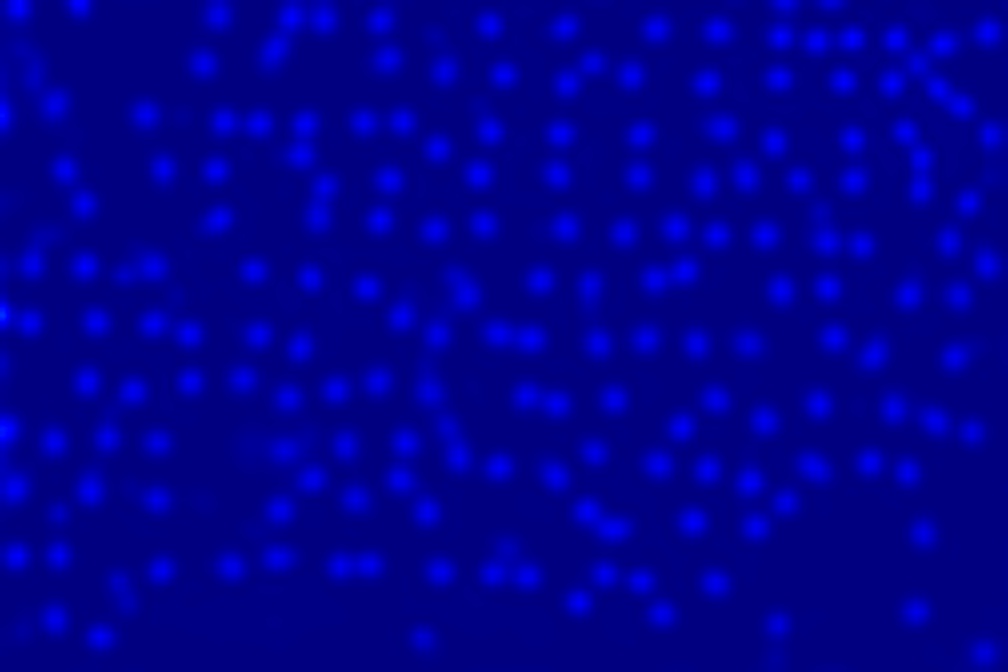}  &
			\includegraphics[height=0.13\linewidth]{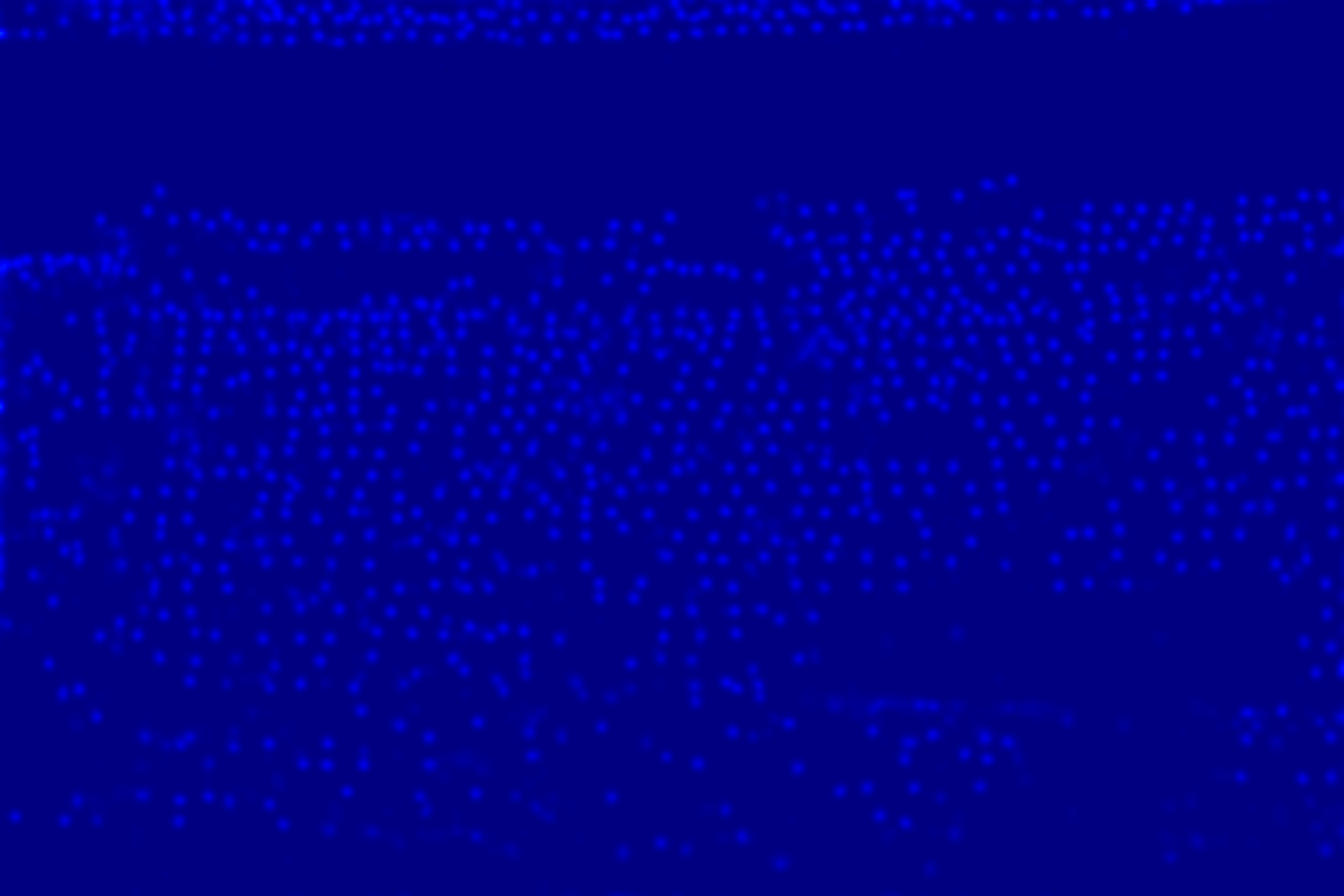}  &
			\includegraphics[height=0.13\linewidth]{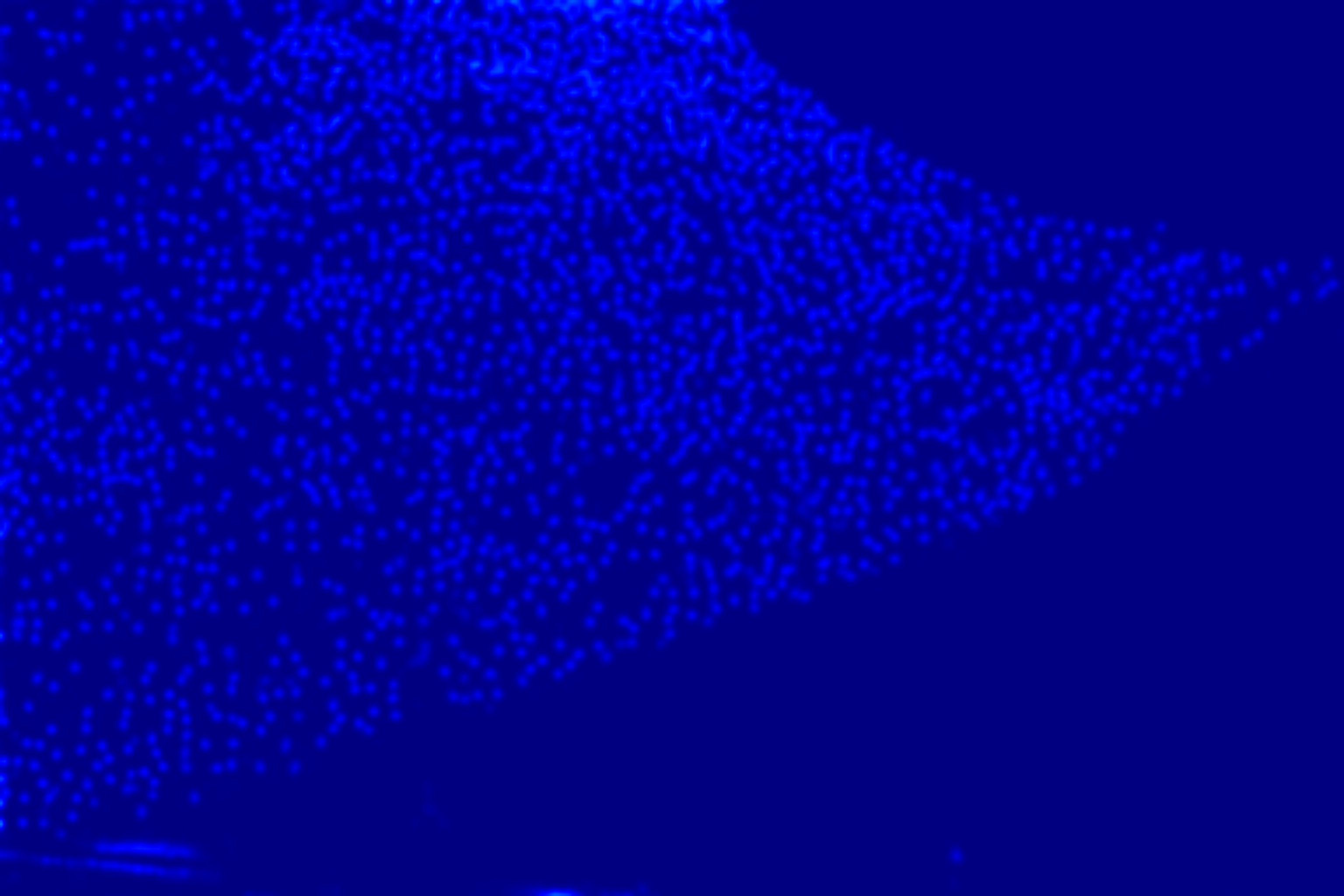}  &
			\includegraphics[height=0.13\linewidth]{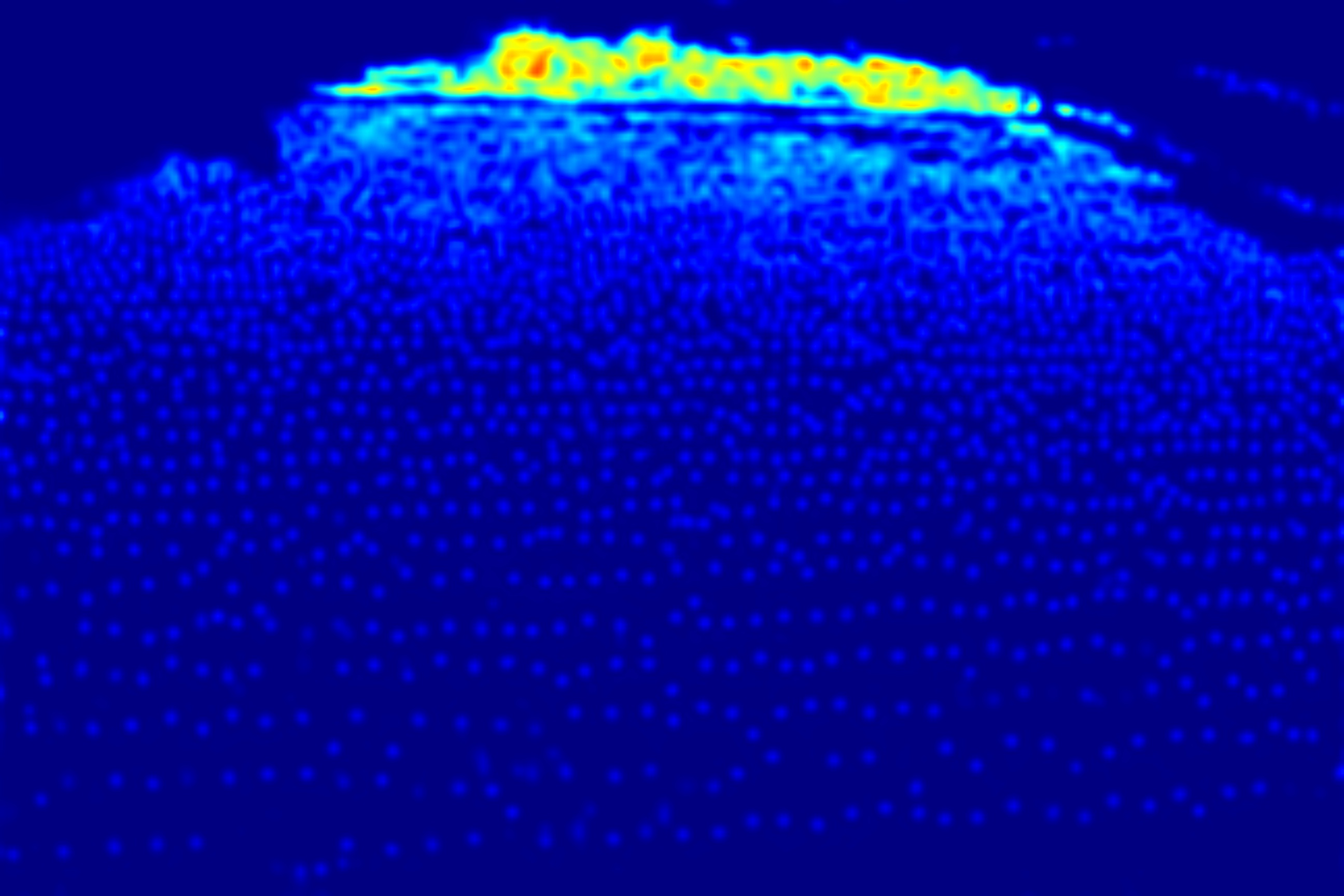}  &
			\includegraphics[height=0.13\linewidth]{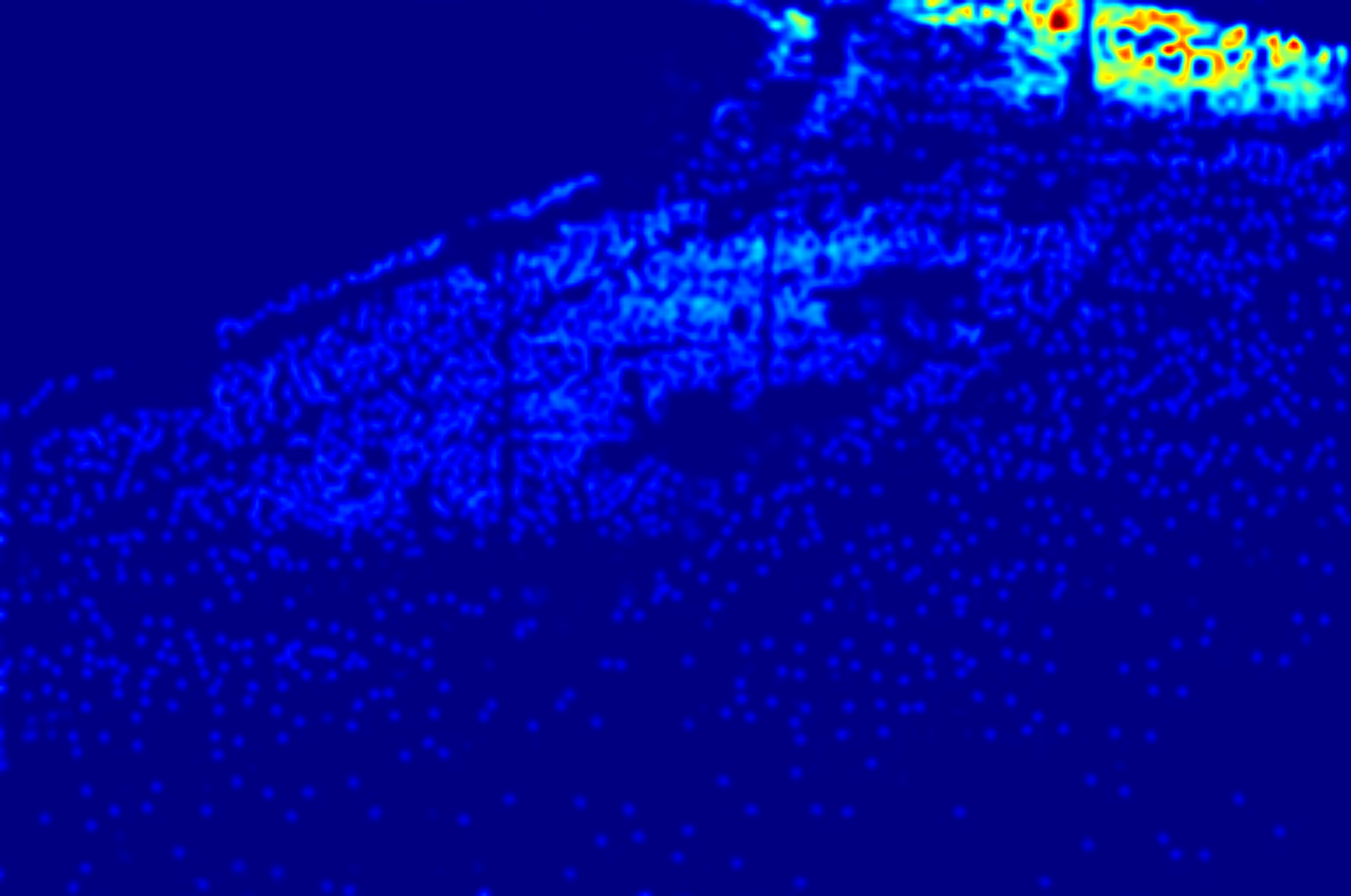}\\

			\footnotesize{baseline: 198.6} & \footnotesize{baseline: 1077.3} & \footnotesize{baseline: 2080.8} & \footnotesize{baseline: 3876.3} & \footnotesize{baseline: 3690.1} \\
			
			\includegraphics[height=0.13\linewidth]{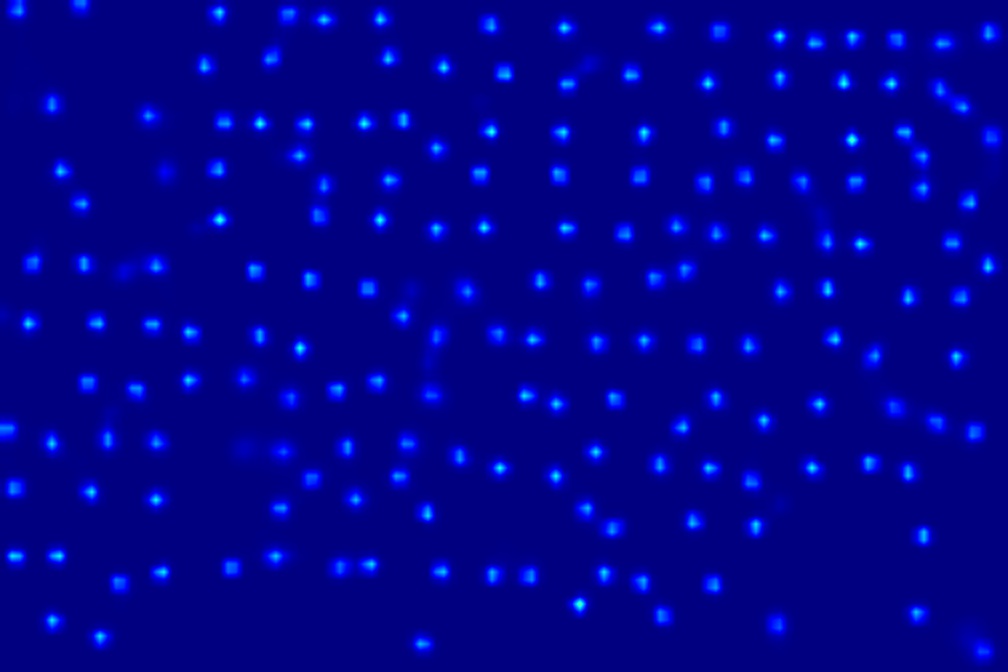}  &
			\includegraphics[height=0.13\linewidth]{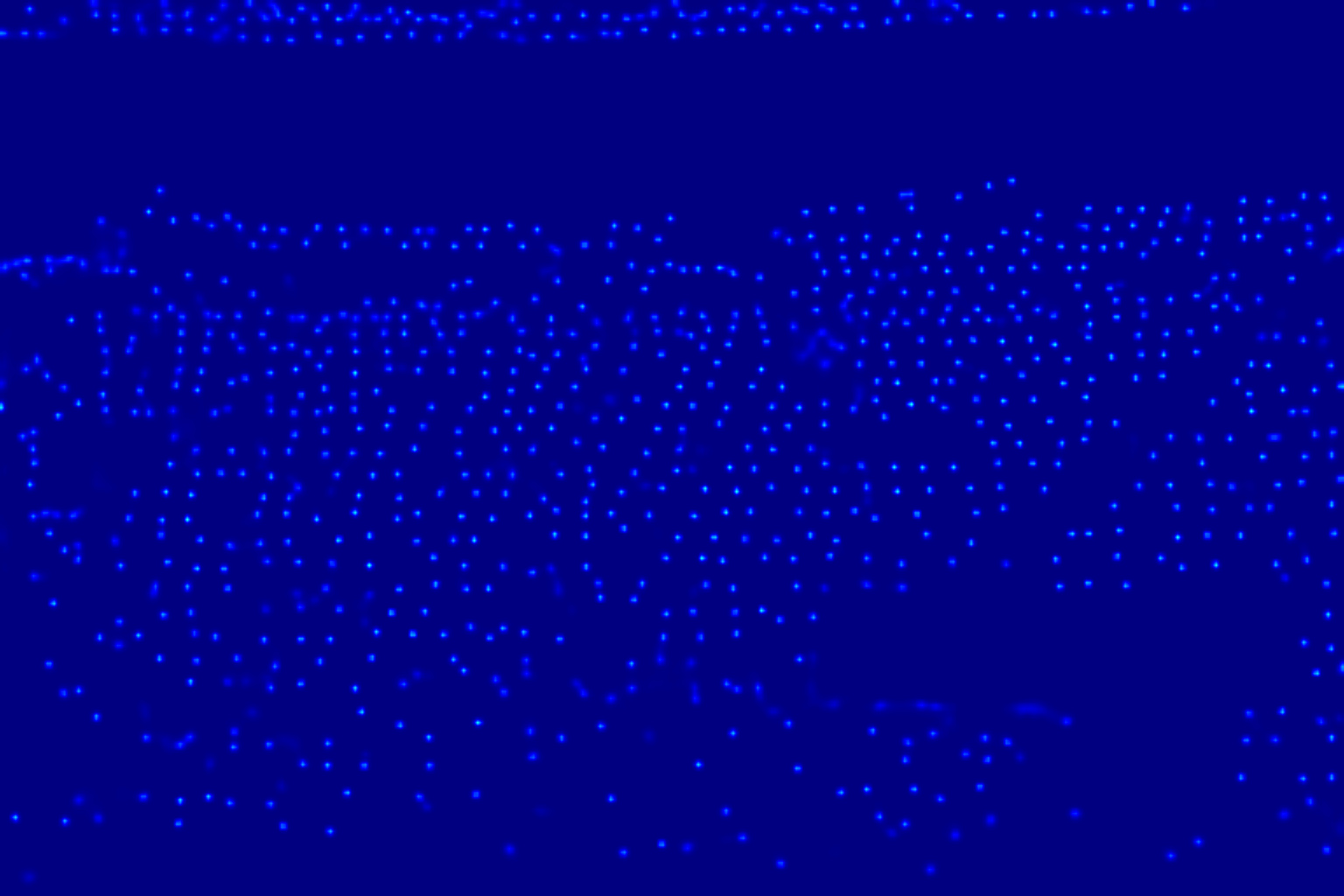}  &
			\includegraphics[height=0.13\linewidth]{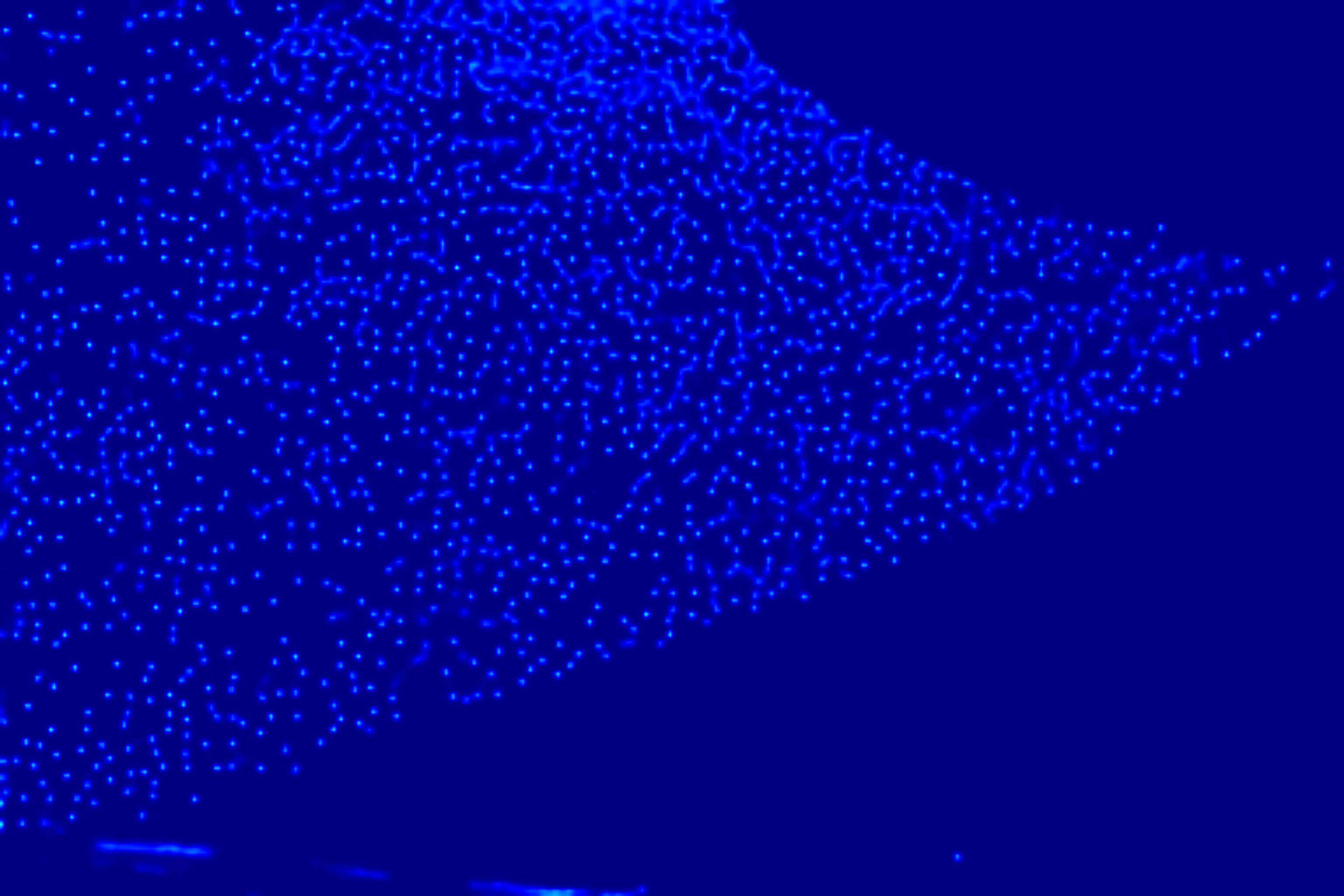}  &
			\includegraphics[height=0.13\linewidth]{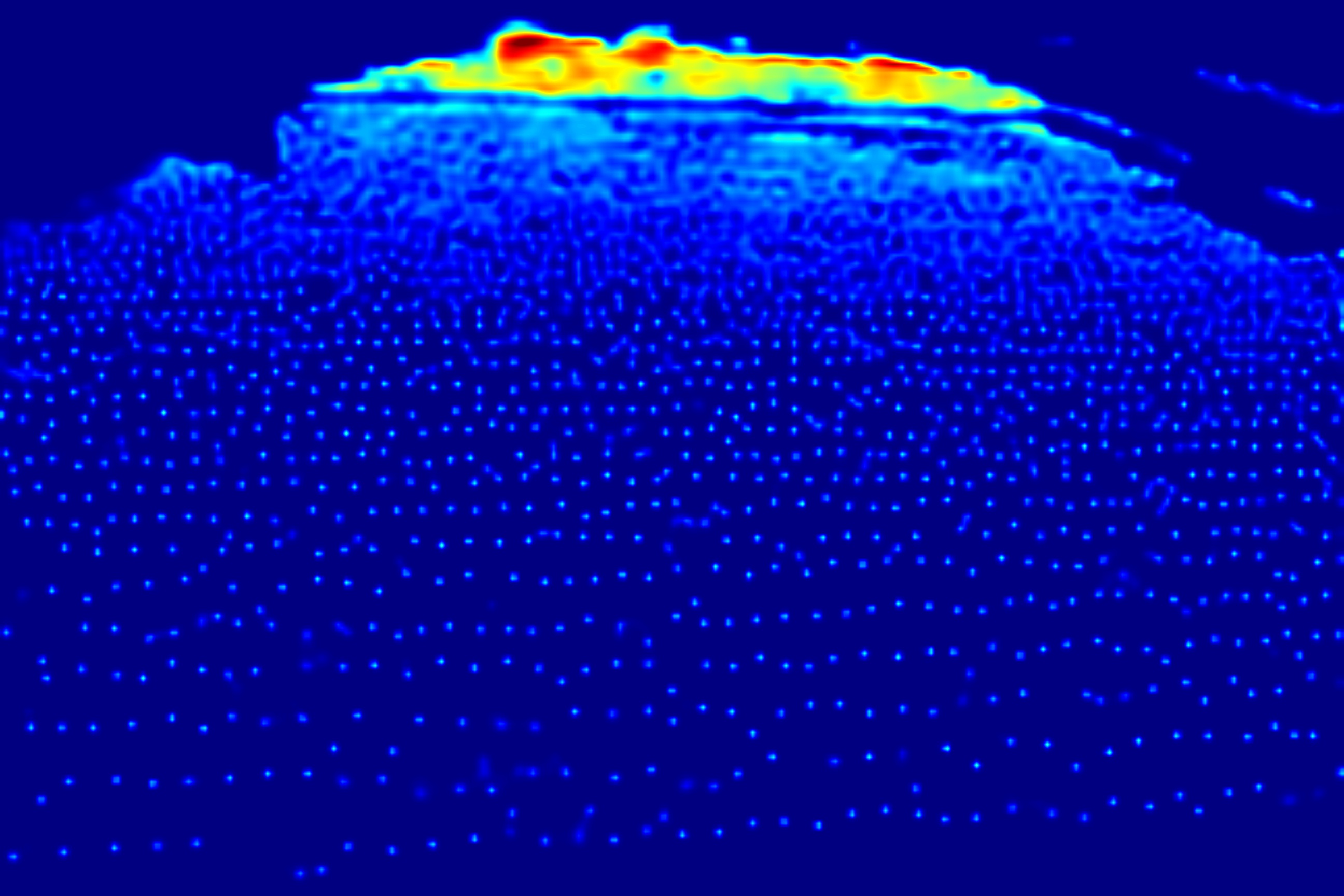}  &
			\includegraphics[height=0.13\linewidth]{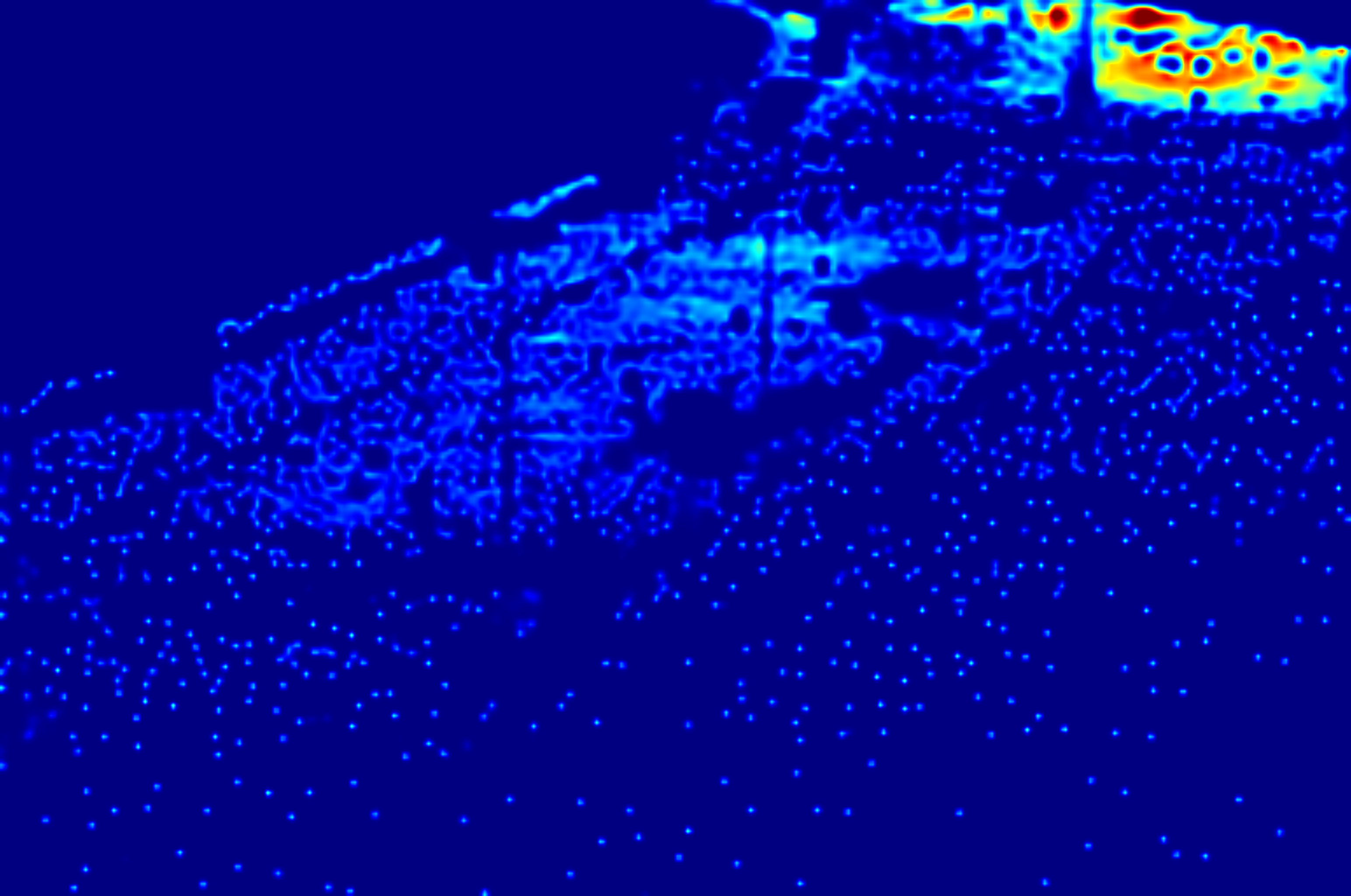}\\
			\footnotesize{S3: 185.3} & \footnotesize{S3: 978.2} & \footnotesize{S3: 2119.1} & \footnotesize{S3: 4010.2} & \footnotesize{S3: 4073.0} \\
		\end{tabular}
		\caption{Visualizations of predicted density maps of the $L_2$ baseline and our proposed semi-balanced Sinkhorn with scale consistency. The first row: input images. The second row: predicted density maps by $L_2$ baseline. The third row: predicted density maps by S3. 
		}
		\label{fig:viz}
	\end{center}
\end{figure*}

\def\arraystretch{1}
\renewcommand{\tabcolsep}{7 pt}{
\begin{table*}[t!]
\footnotesize
	\begin{center}
		\begin{tabular}{c|cc|cc|cc|cc|cc}
			\toprule[1.5pt]
			\multicolumn{1}{c}{Dataset} & \multicolumn{2}{c}{ShanghaiTech A} & \multicolumn{2}{c}{ShanghaiTech B} & \multicolumn{2}{c}{UCF-QNRF} &  \multicolumn{2}{c}{JHU++} & \multicolumn{2}{c}{NWPU} \\
			
			\multicolumn{1}{c}{Method} & \multicolumn{1}{c}{MAE} &
			\multicolumn{1}{c}{MSE} & \multicolumn{1}{c}{MAE} &
			\multicolumn{1}{c}{MSE} & \multicolumn{1}{c}{MAE} &
			\multicolumn{1}{c}{MSE} &  \multicolumn{1}{c}{MAE} &
			\multicolumn{1}{c}{MSE} &  \multicolumn{1}{c}{MAE} &
			\multicolumn{1}{c}{MSE}\\
			\hline
			\hline
			MCNN~\cite{zhang2016single}     & 110.2 & 173.2 & 26.4 & 41.3 & 277 & 426 & 188.9 & 483.4 & 232.5 & 714.6 \\
			CP-CNN~\cite{sindagi2017generating}      & 73.6 & 106.4 & 20.1 & 30.1 & - & - & - & - & - & - \\
			CSRNet~\cite{li2018csrnet}      & 68.2 & 115.0  & 10.6 & 16.0  & - & - & 85.9 & 309.2 & 121.3 & 387.8 \\
			SANet~\cite{cao2018scale}      & 67.0  & 104.5 & 8.4 & 13.6 & - & - & 91.1 & 320.4 & 190.6 & 491.4 \\
			
			CAN~\cite{liu2019context}     & 61.3 & 100.0 & 7.8 & 12.2 & 107.0 & 183.0 & 100.1 & 314.0 & 106.3 & 386.5 \\ 
			
			MBTTBF~\cite{sindagi2019multi}  & 60.2  & 94.1 & 8.0 & 15.5  & 97.5 & 165.2 & 81.8 & 299.1 & - & - \\
			
			BL~\cite{ma2019bayesian}          & 62.8 & 101.8 & 7.7 & 12.7 & 88.7  & 154.8 & 75.0 & 299.9 & 105.4 & 454.2 \\
		    CG-DRCN-CC~\cite{sindagi2020jhu}  & 60.2  & \textbf{94.0} & 7.5 & 12.1  & 95.5 & 164.3 & 71.0 & 278.6 & - & - \\
		    DM-Count~\cite{wang2020distribution} & 59.7 & 95.7 & 7.4 & 11.8 & 85.6 & 148.3 & - & - & 88.4 & 388.6 \\
		    UOT~\cite{ma2021learning} & 58.1 & 95.9 & 6.5 & \textbf{10.2} & 83.3 & 142.3 & 60.5 & 252.7 & 87.8 & 387.5 \\
			\hline	
			$L_2$ Baseline                      & 70.8 & 106.2 & 10.5 & 19.7 & 107.2 & 164.6 & 81.7 & 304.5 & 126.2 & 528.2 \\
			S3 (Ours)                      & \textbf{57.0} & 96.0 & \textbf{6.3} & 10.6 & \textbf{80.6} & \textbf{139.8} & \textbf{59.4} & \textbf{244.0} & \textbf{83.5} & \textbf{346.9} \\
			\bottomrule[1.5pt]
		\end{tabular}
	\end{center}
\caption{Comparisons with the state of the arts on ShanghaiTech, UCF-QNRF, JHU++ and NWPU four crowd benchmarks. $L_2$ baseline and our method are both based on VGG-19.}\label{tab:performance comparison}
\end{table*}}

\subsection{Overall Loss and Optimization}~\label{overall loss and optimization}

{The overall loss of semi-balanced Sinkhorn with scale consistency (S3) is formulated as:
\begin{equation}
\mathcal{L}_{S3}=\mathcal{L}_{smb}+\lambda\mathcal{L}_{sc}=S_\varepsilon^{smb}(\boldsymbol\alpha,\boldsymbol\beta)+\lambda S_\varepsilon(\hat{\boldsymbol\alpha},\boldsymbol{\alpha^\prime}).\label{total loss}
\end{equation}}

To compute the divergence efficiently, we will find the solution from the perspective of dual transform.

Let us first introduce the Fenchel-Legendre conjugate of $\varphi$:
\begin{equation}
\varphi^{*}(z)=\sup_{x}\ zx-\varphi(x)=-\inf_{x}\ \varphi(x)-zx.
\end{equation}

Here we choose Kullback-Leiber as $\varphi$. The divergence functions can be written by:
\begin{equation}
\varphi(x) = \left\{
\begin{aligned}
&x\ln(x)-x+1 &x>0\\
&1 &x=0
\end{aligned}
\right.,\quad \varphi^{*}(z)=e^z -1.
\end{equation}

Kantorovich~\cite{kantorovich1942translocation} proposed the dual transform of Wasserstein distance to relax the hard constraints of the deterministic nature of transportation. The dual form reads as:
\begin{equation}
\begin{aligned}
G_\varepsilon(\boldsymbol\alpha,\boldsymbol\beta)&=\max_{f,g}\ \int_\mathcal{X}{f(x)}{\mathrm{d}\alpha(x)}+\sum_{j=1}^{M}{g(y_j)}{\beta(y_j)}\\
&=\min_{\pi(x,y)}\ W_\varepsilon(\boldsymbol\alpha,\boldsymbol\beta).
\end{aligned}
\label{epsilon wasserstein}
\end{equation}

Based on the Kantorovich theorem, we can then establish the duality of semi-balanced Wasserstein divergence $W_\varepsilon^{smb}(\boldsymbol\alpha,\boldsymbol\beta)$ in Eq.~\ref{sinkhorn} by using Lagrangian term. We derive the formula as:
\begin{small}
\begin{equation}
\begin{aligned}
G_\varepsilon^{smb}(\boldsymbol\alpha,\boldsymbol\beta)&=\sup_{f,g}\ -\int_\mathcal{X} \varphi^{*}(-f(x))\mathrm{d}\alpha(x)+\sum_{j=1}^{M}{g(y_j)}{\beta(y_j)}\\
&-\varepsilon\int_{\mathcal{X}\times\mathcal{Y}}\varphi^{*}(\frac{f(x)+g(y)-c(x,y)}{\varepsilon}){\mathrm{d}\alpha(x)}{\mathrm{d}\beta(y)}.
\end{aligned}
\label{dual transform}
\end{equation}
\end{small}

While optimal solutions have been found, (i.e. $\pi(x,y)$ for the primal, ${f(x)}{g(y)}$ for the dual), there is a relationship between two forms: 
\begin{center}
$\mathrm{d}\pi(x,y)=\exp{\frac{f(x)+g(y)-c(x,y)}{\varepsilon}}\mathrm{d}\alpha(x)\mathrm{d}\beta(y)$.
\end{center}
Therefore, finding the optimal transport map $\pi(x,y)$ is simplified to calculating the optimal dual pair ${f(x)}{g(y)}$, which significantly reduces the number of variables and computational cost. A similar proof can be found in regularized Wasserstein~\cite{kantorovich1942translocation}.

Scaling Algorithm~\cite{sinkhorn1964relationship}~\cite{chizat2018scaling} is an efficient way to compute the optimal dual pair. It gives a positive matrix to iteratively scale the dual vectors by evaluating the primal-dual relationships.

Considering the computational cost, we are inspired by this algorithm and extend it to adjust our problem. When it comes to semi-balanced form, the first order condition of $G_\varepsilon^{smb}$ in Eq.~\ref{dual transform} reads as:
\begin{equation}
\begin{aligned}
&\boldsymbol{f}(x)\big|_{g,\beta}=-\frac{\varepsilon}{1+\varepsilon}\ln(\sum_{j=1}^{M} \exp(\frac{g(y_j)-c(x,y_j)}{\varepsilon})\beta(y_j)),\\ &\boldsymbol{g}(y)\big|_{f,\alpha}=-\varepsilon\ln(\int_{\mathcal{X}} \exp(\frac{f(x)-c(x,y)}{\varepsilon})\mathrm{d}\alpha(x)).
\end{aligned}
\label{fgiter}
\end{equation}
Here, $f(x)$ is in the \emph{unbalanced} form (of dual vectors) while $g(y)$ is in the \emph{balanced} one.


Thus, for balanced self-correcting term $W_{\varepsilon}(\boldsymbol\alpha,\boldsymbol\alpha)$, the symmetric optimal iteration of dual vector $p(x)$ is similar to balanced dual vector $g(y)$:
\begin{equation}
\boldsymbol{p}(x)\big|_{p,\alpha}=-\varepsilon\ln(\int_{\mathcal{X}} \exp(\frac{p(x)-c(x,x)}{\varepsilon})\mathrm{d}\alpha(x)).
\label{piter}
\end{equation}

Given the density measure $\boldsymbol\alpha$ and the annotated measure $\boldsymbol\beta$, if there is no person in the image, we will only regress $\boldsymbol\alpha$ to the zero vector. Otherwise, the cross correlation and self-correcting dual vectors will be computed by the Scaling Algorithms. The semi-balanced Sinkhorn distance, which is used as counting loss $\mathcal{L}_{smb}$ in Eq.~\ref{total loss}, is calculated as: 
\begin{equation}
\begin{aligned}
S_\varepsilon^{smb}(\boldsymbol\alpha,\boldsymbol\beta)=\int_\mathcal{X}(-\varphi^{*}(-f(x))-p(x))\mathrm{d}\alpha(x)\\
+\sum_{j=1}^{M}{g(y_j)}{\beta(y_j)}
+\frac{\varepsilon^2}{2}(m(\boldsymbol\alpha)-m(\boldsymbol\beta))^{2}.
\end{aligned}
\label{semibalanced optimization}
\end{equation}
{The computation of scale consistency loss $\mathcal{L}_{sc}$ can be performed analogously, by replacing the \emph{unbalanced} dual vector $f(x)$ to its \emph{balanced} form and substituting it into Eqs.~\ref{epsilon sinkhorn} and~\ref{epsilon wasserstein}.}
More details of optimization are summarized in Algorithm~\ref{algorithm}.

\begin{algorithm}[t]
\caption{{S3 Optimization}}\label{algorithm}
\KwIn{Density regressor $R$ with parameter $\theta$, input image $I$, ground truth measure $\boldsymbol{\beta}$, scale transform $Sc$}
\KwOut{density regressor $R$ with optimized parameter $\hat{\theta}$}
Initialize $\theta^{(1)}$;\\
\For{epoch $t=1,\ldots,T$}{
$\boldsymbol{\alpha}^{(t)} = R_{\theta^{(t)}}(I)$; \\
$\boldsymbol{\hat{\alpha}}^{(t)} = R_{\theta^{(t)}}(Sc(I))$; \\
Initialize $i=1, f^{(t,i)} = \mathbf{0}_{N^{t}}, g^{(t,i)} = \mathbf{0}_{M^{t}}$; \\ 
\tcp{$\mathbf{0}_{N^{t}}$, $\mathbf{0}_{M^{t}}$ are zero vectors}
\Repeat{convergence}{
$f^{(t,i+1)} = \boldsymbol{f}(x)\big|_{g^{(t,i)}, \boldsymbol\beta}$; \\ 
$g^{(t,i+1)} = \boldsymbol{g}(y)\big|_{f^{(t,i+1)}, \boldsymbol{\alpha}^{(t)}}$; \ (Eq.~\ref{fgiter})\\ 
$i=i+1$;
}
Perform iterations in steps 6-10 for $p^{(t)}$ until convergence;  (Eq.~\ref{piter}) \\
$\mathcal{L}_{smb}^{(t)}=S_{smb}(f^{(t)}, g^{(t)}, p^{(t)}, \boldsymbol{\alpha}^{(t)}, \boldsymbol{\beta})$; \ (Eq.~\ref{semibalanced optimization})\\
$\mathcal{L}_{sc}^{(t)}=S_{\varepsilon}(\boldsymbol{\hat{\alpha}}^{(t)}, Sc(\boldsymbol{\alpha}^{(t)}))$; \ (Eq.~\ref{sc loss})\\
$\mathcal{L}_{S3}^{(t)}=\mathcal{L}_{smb}^{(t)}+\lambda\mathcal{L}_{sc}^{(t)}$; \ (Eq.~\ref{total loss}) \\
Minimize $\mathcal{L}_{S3}^{(t)}$ by optimizing $\theta$; \\
Update $\theta^{(t+1)}$ using Adam;
}
\textbf{Return}  $\hat{\theta}=\theta^{(t+1)}$.
\end{algorithm}

\section{Experimental Results}

\subsection{Implementation Details and Datasets}

We have conducted extensive experiments on four largest crowd counting benchmarks which are widely used in recent papers. VGG-19 has been adopted as our network structure and the whole code is implemented by Pytorch. The influence of different key parameters will be detailed in Section~\ref{ablation study}.

\noindent\textbf{ShanghaiTech}~\cite{zhang2016single} includes Part A and Part B. In Part A, there are 482 images with 244,167 annotated points. 300 images are divided for training and the remaining 182 images are for testing. In Part B, there are 716 images with 88,498 annotated points. 400 images are divided for training and the remaining 316 images are for testing.

\noindent\textbf{UCF-QNRF}~\cite{idrees2018composition} includes 1,535 images with 1.25 million annotated points. It has a wide range of people count and images with high resolutions. The training set contains 1,201 images and the testing set includes the rest 334 images.

\noindent\textbf{JHU-CROWD++}~\cite{sindagi2020jhu} includes 4,372 images with 1.51 million annotated points. 2,272 images are chosen for training; 500 images are for validation; and the rest 1,600 images are for testing. Compared to others, JHU-CROWD++ contains diverse scenarios and is collected under different environmental conditions of weather and illumination.

\noindent\textbf{NWPU-CROWD}~\cite{wang2020nwpu} contains 5,109 images with 2.13 million annotated points. 3,109 images are divided into training set; 500 images are in validation set; and the remaining 1,500 images are in testing. Images in NWPU-CROWD are in largely various density and illumination scenes.


\subsection{Comparisons with the State of the Arts}

In this section, we evaluate our results on above four datasets and list ten recent state-of-the-arts methods for comparison. 

The error of counting task is calculated by two commonly used metrics, Mean Absolute Error (MAE) and Mean Squared Error (MSE). The lower of both means the better performance.~\cite{zhang2016single} 

Visualizations of the predicted density maps are shown in Figure ~\ref{fig:viz}. {The outputs of the proposed method appear sharp and are closed to the locations of crowds.} 


\subsection{Quantitative Results Analysis}

Counting accuracy of our method is presented in Table~\ref{tab:performance comparison} and our proposed semi-balanced Sinkhorn with scale consistency is denoted as S3. We perform lower MAE and MSE in this task, which proves the improvements and merits of our method. Highlights are summarized as follows:
\begin{itemize}
    \item S3 significantly improves the counting accuracy on ShanghaiTech B, UCF-QNRF, JHU++ and NWPU crowd datasets. Especially, on QNRF, S3 improves MAE and MSE values of UOT~\cite{ma2021learning} from 83.3 to 80.6 and from 142.3 to 139.8, respectively.
    
    \item Without any external structures or detection methods, S3 significantly improves the performance of traditional pseudo-map-regressive $L_2$ baseline on all four benchmarks.
\end{itemize}




\subsection{Ablation Study}~\label{ablation study}


In this section, we hold ablation experiments to study the influence of loss terms and key parameters.

\noindent
\textbf{Contribution of loss terms.} The overall loss is the combination of a semi-balanced Sinkhorn counting loss and a scale consistency loss. We study the contribution of each loss term in Table~\ref{tab:contribution} and quantitative results verify the effectiveness of our proposed losses. 

\begin{itemize}
    \item The semi-balanced Sinkhorn counting loss $\mathcal{L}_{smb}$ remarkably promotes traditional $L2$ loss in counting accuracy. MAE and MSE are improved by \textbf{23.2} and \textbf{18.8} respectively. Then the combination of scale consistency $\mathcal{L}_{sc}$ stabilizes our model and further reduces the error by 3.4 and 6.0.
\end{itemize}
\begin{table}[H]
\footnotesize
\begin{center}
\begin{tabular}{c|cccc}
\toprule[1pt]
\multicolumn{1}{c}{Loss terms} & $L_2$ & $\mathcal{L}_{WD}$ & $\mathcal{L}_{smb}$ & S3 \\
\hline
\small
MAE & 107.2 & 98.4 & 84.0 & 80.6 \\
MSE & 164.6 & 169.6 & 145.8 & 139.8\\
\toprule[1pt]
\end{tabular}
\end{center}
\caption{Comparison of loss terms on UCF-QNRF.}
\label{tab:contribution}
\end{table}

\begin{figure}[t]
\begin{center}
    \includegraphics[width=0.5\textwidth]{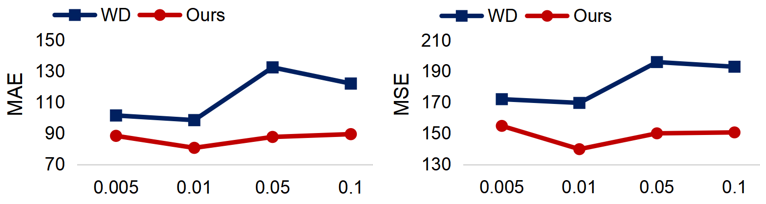}
\end{center}
\caption{The influence of different smooth parameter $\varepsilon$ on UCF-QNRF. Our proposed S3 performs more stable than traditional Wasserstein.}
\label{fig:smooth}
\end{figure}

\noindent
\textbf{Different smooth parameter $\boldsymbol{\varepsilon}$.} $\varepsilon$ controls the level of regularization, causing entropic bias in traditional Wasserstein distance (WD). We compare the fluctuations for using different $\varepsilon$ in WD and in our proposed semi-balanced Sinkhron with scale consistency (S3) in Figure~\ref{fig:smooth}. 

\begin{itemize}
    
    \item Compared to WD, S3 outperforms consistently and is more stable. MAE and MSE of WD vary from 98.4 to 132.3 and from 169.6 to 195.9, respectively. On contrast, those of S3 vary from 80.6 to 89.4 and from 139.8 to 150. The quantitative results can justify that our method is able to eliminate the entropic bias to some extent.
\end{itemize}

\section{Conclusions}

{In this paper, we propose a novel measure matching based crowd counting approach, termed semi-balanced Sinkhorn with scale consistency. S3 has several advantages. 1) It avoids generating pseudo density maps with erroneous size assumptions, by allowing to use ground truth points as supervised signal. 2) The semi-balanced Sinkhorn addresses the entropic bias and amount constraints existing in other traditional measure divergences. 3) The Sinkhorn scale consistency loss stabilizes our method under the scenarios with various crowd scales. 4) The proposed pipeline  works only in the learning stage and thus doesn't bring any extra computational burdens in prediction. In future, we plan to extend the proposed method to video based crowd counting.}


\section*{Acknowledgements}
	This work is funded by National Key Research and Development Project of China under Grant No.  2019YFB1312000 and 2020AAA0105600, National Natural Science Foundation of China under Grant No. 62076195, 62006183, U20B2052, and 62006182, and by China Postdoctoral Science Foundation under Grant No. 2020M683489.


\bibliographystyle{named}
\bibliography{ijcai21}

\end{document}